\def\year{2022}\relax
\documentclass[letterpaper]{IEEEtran} 
\usepackage{times}  
\usepackage{helvet}  
\usepackage{courier}  
\usepackage[hyphens]{url}  
\usepackage{graphicx} 
\usepackage{caption} 
\DeclareCaptionStyle{ruled}{labelfont=normalfont,labelsep=colon,strut=off} 
\usepackage{algpseudocode}                                 
\usepackage{algorithm}
\usepackage{subfig}
\usepackage{newfloat}
\usepackage{listings}
\floatstyle{ruled}
\newfloat{listing}{tb}{lst}{}
\floatname{listing}{Listing}

\graphicspath{{./figs/}}

\usepackage{amsmath,amsfonts,bm}

\def\eqref#1{equation~\ref{#1}}

\def\1{\bm{1}}

\def\vb{{\bm{b}}}

\def\vh{{\bm{h}}}

\def\vm{{\bm{m}}}

\def\vx{{\bm{x}}}
\def\vy{{\bm{y}}}

\def\mA{{\bm{A}}}

\def\mC{{\bm{C}}}

\DeclareMathAlphabet{\mathsfit}{\encodingdefault}{\sfdefault}{m}{sl}
\SetMathAlphabet{\mathsfit}{bold}{\encodingdefault}{\sfdefault}{bx}{n}

\def\gA{{\mathcal{A}}}
\def\gB{{\mathcal{B}}}

\def\gE{{\mathcal{E}}}

\def\gG{{\mathcal{G}}}

\def\gN{{\mathcal{N}}}

\def\gV{{\mathcal{V}}}

\def\chi{{\mathcal{X}}}

\def\sX{{\mathbb{X}}}

\newcommand{\R}{\mathbb{R}}

\usepackage{amsmath,amsfonts,amssymb,amsthm}
\usepackage{tikz}
\usepackage{pgfplots}
\usepackage{pgfplotstable}
\pgfplotsset{compat=newest}
\usepackage{filecontents}
\usepackage{booktabs}             
\usepackage{tabularx}
\usepackage{multirow}
\usepackage{cleveref}
\usepackage{xcolor}
\definecolor{Constrastive-Distillation}{RGB}{40,82,122}
\definecolor{w/o distillation}{RGB}{204,86,30}
\definecolor{DELTA}{RGB}{244,247,197}
\definecolor{mygreen}{RGB}{134, 182, 54}
\definecolor{myblue}{RGB}{51, 51, 255}
\definecolor{myyellow}{RGB}{255, 255, 191}
\definecolor{myorange}{RGB}{244, 106, 18}
\definecolor{mygray}{RGB}{102, 102, 102}
\definecolor{mymiddle}{RGB}{167, 44, 167}
\definecolor{mylight}{RGB}{191, 191, 255}
\definecolor{mydark}{RGB}{114, 44, 114}
\definecolor{mypink}{RGB}{255,171,164}  
\definecolor{_orange}{RGB}{230,170,117}
\definecolor{_blue}{RGB}{117,177,230}
\definecolor{_gray}{RGB}{119,136,153}
\definecolor{lightblue}{RGB}{221,239,250}  
\newcommand{\model}{GDN }
\newcommand{\modelnospace}{GDN}

\newtheorem{mydefinition}{\textbf{Definition}}
\newtheorem{mytheorem}{\textbf{Theorem}}
\newtheorem{mycorollary}{\textbf{Corollary}}

\newtheorem{myprop}{Property}
\begin{document}
\title{Rethinking Graph Neural Networks for the Graph Coloring Problem}

\author{
    \IEEEauthorblockN{
        Wei Li\IEEEauthorrefmark{1}, Ruxuan Li\IEEEauthorrefmark{2}, Yuzhe Ma\IEEEauthorrefmark{3}, Siu On Chan\IEEEauthorrefmark{2}, David Pan\IEEEauthorrefmark{4}, Bei Yu\IEEEauthorrefmark{2}
    } \\
    \IEEEauthorblockA{
        \textit{\IEEEauthorrefmark{1}Carnegie Mellon University \qquad \IEEEauthorrefmark{2}The Chinese University of Hong Kong \\
        \IEEEauthorrefmark{3}The Hong Kong University of Science and Technology (Guangzhou) \\ \IEEEauthorrefmark{4}The University of Texas at Austin}
    }
}

\maketitle

\begin{abstract}
    Graph coloring, a classical NP-hard problem, is the problem of assigning connected nodes as different colors as possible.
    In this work, we aim to solve the coloring problem by graph neural networks (GNNs).
    However, we observe that state-of-the-art GNNs are less successful in the graph coloring problem.
    We analyze the reasons from two perspectives.
    First, most GNNs fail to generalize the task under homophily to heterophily, i.e., graphs where connected nodes are assigned different features. 
    Second, GNNs are bounded by the network depth, making them possible to be a local method, which has been demonstrated to be non-optimal in Maximum Independent Set (MIS) problem.
    In this paper, we focus on the \textit{aggregation-combine GNNs} (AC-GNNs), a popular class of GNNs.
    Instead of learning when AC-GNNs assign local equivalent node pairs to the same node embedding, which is designed for the task under homophily,
    we study the power of a GNN in the coloring problem by analyzing its ability to assign nodes different colors.
    Through the analysis, we find some specific settings/architectures that may harm the performance.
    Furthermore, we demonstrate the non-optimality of AC-GNNs due to their local property, and prove the positive correlation between model depth and its coloring power.
    Following the discussions above, we summarize a series of rules that make a GNN powerful in the coloring problem. 
    Then, we propose a simple AC-GNN variation based on these rules.
    We empirically validate our theoretical findings and demonstrate that our simple model substantially outperforms state-of-the-art heuristic algorithms in both quality and runtime.
\end{abstract}

\section{Introduction}
Graph neural networks (GNNs) have shown overwhelming success in various fields, such as molecules, social networks, and web pages\cite{hamilton2017representation}.
The main idea behind GNNs is a neighborhood aggregation scheme (or called message passing),
where each node aggregates feature vectors from its neighbors and combines them with its own feature vector to produce a new one.
GNNs following such a scheme are called aggregation-combination GNNs (AC-GNNs) \cite{barcelo2019logical}.
After finite iterations of aggregation and combination, the corresponding feature vector of each node is called node embedding to represent the node.


In this work, we try to study the performance of GNNs for the coloring problem because
1) Graph coloring and its variations have a great demand in industry. 
while the target graph size is exploding nowadays.
For example, the netlist graph in a commercial chip contains millions of gate nodes;
the size of the user graph in the Internet company is also at a million level.
These graphs are too large to be processed by transitional algorithms within an acceptable time, and 
therefore motivate us to apply the power of highly parallelable GNNs.
2) 
However, in our motivating experiment, most of existing GNNs even cannot beat the simplest heuristic algorithm.
Therefore, we try to find the reason for the low performance
so that we can provide some theoretical guidance on a powerful GNN for the coloring problem.
3)
Graph problems under heterophily are the ones where connected nodes are expected to have \textit{different} labels/features/colors instead of a similar one (homophily),
and graph coloring problem is the most representative task under heterophily.
Although some rules are shown to be able to enhance the expressive power of GNNs in previous work,
whether these rules still hold under heterophily is still an open question.
For example, in previous study under homophily, 
deeper GNNs often suffer from over-smooth problem (Zhao and Akoglu 2019) and it is believed that deepening GNNs do not improve (or sometimes worsen) their performance (Oono and Suzuki 2019).
We will try to find the answer during our explorations in the coloring problem.

\begin{figure*}[tbh!]
	\centering
    \subfloat[]{\includegraphics[width=.65\textwidth]{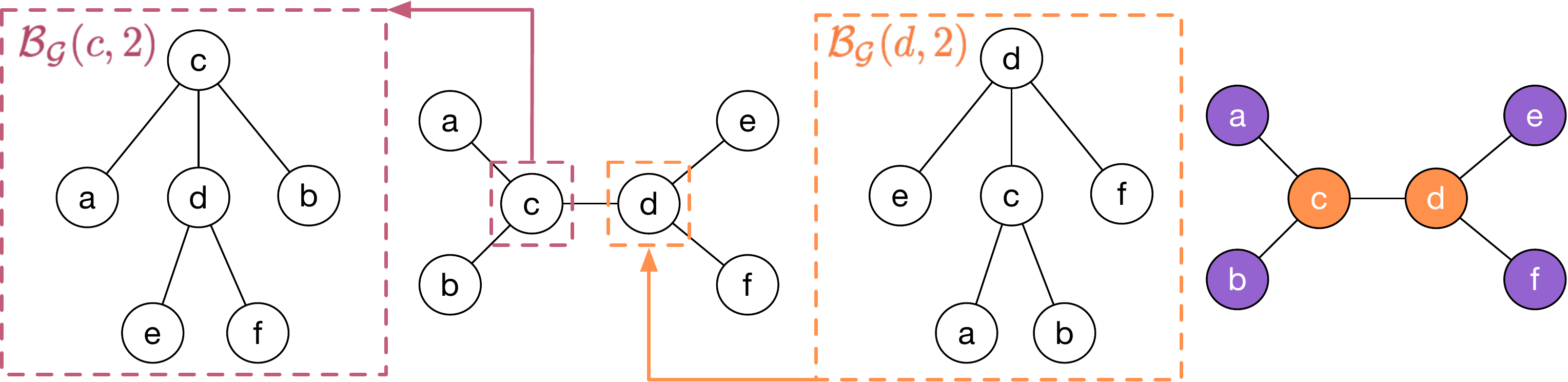} \label{fig:acgnn}} \hspace{.01\textwidth}
    \subfloat[]{\includegraphics[width=.32\textwidth]{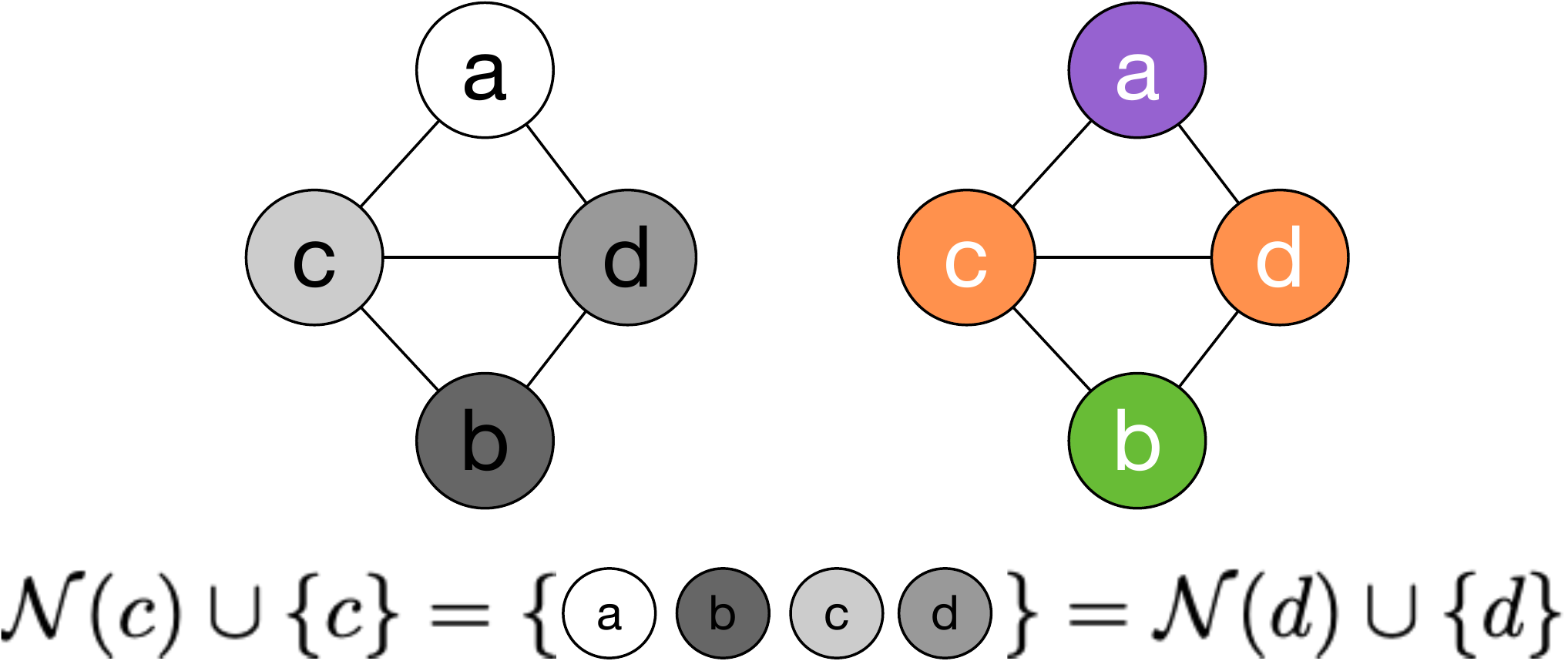}  \label{fig:intgnn}}  
    \caption{
        Examples of graph structures in which some AC-GNNs fail to discriminate the equivalent node pair $\{c,d\}$.
        (a) left: the input graph with the same node attribute; right: the coloring results by the most powerful AC-GNN.
        (b) left: the input graph with different node attributes (represented by the gray scale); right: the coloring results by the most powerful \textit{integrated} AC-GNN.
        The aggregation for any \textit{integrated} AC-GNN in both $c$ and $d$ are the same since $\gN(c) \cup \{c\}= \gN(d) \cup \{d\}$.
        Here, the most powerful is an ideal object, and refers to a virtual integrated AC-GNN that has the most powerful ability to assign node pairs to different colors.
    }
	\label{fig:color}
\end{figure*}

We study the problem by investigating the \textit{power} of GNNs {for the graph coloring problem}.
Some recent works \cite{loukas2019graph,barcelo2019logical,xu2018powerful,gama2020stability,loukas2020hard,du2019graph} 
study the power of a GNN by analyzing when a GNN maps two nodes to the same node embedding.
In their study, a maximally powerful GNN with depth $L$ should map two $L$-local equivalent nodes to the same node embedding \cite{xu2018powerful,loukas2019graph}.
However, when applied in the graph coloring problem, such a definition raises some problems.
First, the coloring task is not under homophily but \textit{heterophily}.
Therefore, two local equivalent nodes are not necessarily assigned the same node embedding.
One example can be found in Figure \ref{fig:acgnn}, where the node pair $\{c,d\}$ is local equivalent but should be assigned different colors to avoid the conflict.
Second, every AC-GNN is bounded by its depth $L$. Therefore, the maximally powerful GNN is identified by $L$-local equivalence instead of a \textit{global} equivalence.
This constraint makes an AC-GNN possible to be a \textit{local method}, which has been demonstrated to be non-optimal in many NP-hard problems such as MIS \cite{rahman2017local,gamarnik2014limits}.

Motivated by these limitations, we define the power of AC-GNNs in the coloring problem as its ability to assign nodes different colors.
We then observe and theoretically prove a set of conditions that may harm/contribute to the coloring performance.
Based on these observations, we develop a series of rules to design powerful AC-GNNs \textit{specifically for the graph coloring problem}.
We make the following contributions:
(1): We show that AC-GNNs cannot be optimal in the coloring problem and demonstrate the positive correlation between model depth and its power in the coloring problem.
    (2) We give simple but effective rules about the GNN architecture for the coloring problem, some of which are contrary to previous research under homophily.
    (3) Combining these rules, we develop a simple GNN-based approach by un-supervised learning.
    (4) We validate our findings by extensive empirical evaluation including three datasets from different subjects.
    Our method shows substantially superior performance compared with other existing AC-GNN variations and even outperforms state-of-the-art heuristic algorithms with a significant efficiency improvement.

\section{Preliminaries}
\paragraph{Graph Terminology}
\label{appendix:term}
Here we list the following graph theoretic terms encountered in our work.
Let $\gG = (\gV,\gE)$ and $\gG' = (\gV',\gE')$ be graphs on vertex set $\gV$ and $\gV'$, we define
\begin{itemize}
    \item \textit{isomorphism}: we say that a bijection $\pi: \gV \rightarrow \gV'$ is an \textit{isomorphism} if any two vertices $u,v \in \gV$ are adjacent in $\gG$
    if and only if $\pi(u),\pi(v) \in \gV'$ are adjacent in $\gG'$, i.e., $\{u,v\} \in \gE $ iff $\{\pi(u),\pi(v)\} \in \gE' $.
    \item \textit{isomorphic nodes}: If there exists the isomorphism between $\gG$ and $\gG'$, we say that $\gG$ and $\gG'$ are \textit{isomorphic}.
    \item \textit{automorphism}: When $\pi$ is an isomorphism of a vertex set onto itself, i.e., $\gV = \gV'$,
    $\pi$ is called an \textit{automorphism} of $\gG$.
    \item \textit{topologically equivalent}: We say that the node pair $\{u,v\}$ is \textit{topologically equivalent} if there is an automorphism mapping one to the other, i.e., $v = \pi(u)$.
    \item \textit{equivalent}: $\{u,v\}$ is \textit{equivalent} if it is topologically equivalent by $\pi$ and $\vx_w = \vx_{\pi(w)}$ holds for every $w \in \gV$,
    where $\vx_w$ is the node attribute of node $w$.
    \item \textit{$r$-local topologically equivalent}: 
    The node pair $\{u,v\}$ is \textit{$r$-local topologically equivalent}
    if $\pi_r$ is an isomorphism from $\gB_{\gG}(u,r)$ to $\gB_{\gG}(v,r)$.
    \item \textit{$r$-local equivalent}: $\{u,v\}$ is \textit{$r$-local equivalent} if it is $r$-local topologically equivalent by $\pi_r$ and $\vx_w = \vx_{\pi_r(w)}$ holds for every $w \in \gB_{\gG}(u,r)$.
    \item \textit{$r$-local isomorphism}:
    A bijection $\pi_r$ is an \textit{$r$-local isomorphism} that maps $u$ to $v$ if $\pi_r$ is an isomorphism that maps $\gB_{\gG}(u,r)$ to $\gB_{\gG}(v,r)$.
    \item \textit{other local graph terminology}: For every positive integer $r$ and every node $u \in \gV$, we define $\gB_{\gG}(u,r)$ as the subgraph of $\gG$ induced by node $u$ with distance at most $r$ from $u$.
\end{itemize}

One example is given in Figure \ref{fig:acgnn}.

\paragraph{Graph coloring.}
Let $k$ be the number of available colors, $\gG = (\gV,\gE)$ be the input graph and each vertex $v \in \gV$ be associated with an attribute $\vx_v$,
a coloring function $f_k: (v, \gG, \vx_v) \rightarrow  \{1,...,k\}$ returns a color of $v$ indexed by $col_v \in \{1,...,k\}$.
In the following pages, $k$ follows the same definition if not specified and $f(\gG)$ represents the coloring solution on $\gG$ by $f$ for simplification.
Given a graph $\gG$ colored by $f_k$, a \textit{conflict function} $c: (u,v,f_k) \rightarrow \{0,1\}$ is used to measure the performance of $f_k$ on $\gG$.
Specifically, $c(u,v,f_k) = 1$ when $u$ and $v$ are connected and assigned the same color:
\begin{equation}
    \label{eq:conflict}
    c(u,v,f_k) =
\begin{cases}
    1,&  \text{if } \ f_k(v,\gG,\vx_v) = f_k(u,\gG,\vx_u) \\ & \ \text{and} \ \{u,v\}\in \gE;\\
    0,                & \text{otherwise}.
\end{cases}
\end{equation}
The edge $e = \{u,v\}$ is called a \textit{conflict} if $c(u,v,f) = 1$.
The objective of the graph coloring problem is widely formulated in two ways:
1) \textit{$k$-coloring problem}: Given $k$, minimize the number of conflicts as in \Cref{eq:min_cost};
2) Given a conflict constraint $c_{max}$, minimize the number of used colors as in \Cref{eq:min_k}.

    \begin{equation}
        \label{eq:min_cost}
        \min \sum_{\{u,v\}\in \gE}c(u,v,f_k).
    \end{equation}
\begin{minipage}{\linewidth}
    \begin{equation}
        \label{eq:min_k}
        \min k, \ \text{s.t.}~\sum_{\{u,v\}\in \gE}c(u,v,f_k) \leq c_{max}.
    \end{equation}
\end{minipage}

When we set $c_{max}$ as 0, i.e., no conflict is introduced by $f_k$,
we refer the obtained minimum color number as the \textit{chromatic number} of $\gG$, which is often represented by $\chi$.
The corresponding coloring function $f_\chi$ is called an \textit{optimal} function.


\paragraph{Graph neural networks (GNNs).}
GNNs are to learn the node embeddings or graph embedding based on the graph $\gG = (\gV,\gE)$ and node features $\{\vx_v: v\in \gV \}$.
We follow the same notations in \cite{barcelo2019logical} to formally define the basics for GNNs.
Let $\{\text{AGG}^{(i)}\}_{i=1}^L$ and $\{\text{COM}^{(i)}\}_{i=1}^L$ be two sets of \textit{aggregation} and \textit{combination} functions.
An \textit{aggregation-combine GNN} (AC-GNN) computes the feature vectors $\vh_v^{(i)}$ for every node $v\in \gV$ by:
\begin{align}
    \vh_v^{(i)} = \text{COM}^{(i)}(\vh_v^{(i-1)}, \text{AGG}^{(i)}(\{\vh_u^{(i-1)}: u\in \gN(v) \})),
   \end{align}
where $\gN(v)$ denotes the {neighborhood} of $v$, i.e., $\gN(v) = \{u: \{u,v\} \in  \gE \}$ and $\vh_v^{(0)}$ is the node attribute $\vx_v$.
Finally, each node $v$ is classified by a node classification $CLS(\cdot)$ applied to the node embedding $\vh_v^{(L)}$.
When the AC-GNN is used for the graph coloring problem, $CLS(\cdot)$ returns a $col_v \in \{1,...,k\}$.
Then, an AC-GNN $\gA$ with $L$ layers is also called $L$-AC-GNN and defined as $\gA = (\{\text{AGG}^{(i)}\}_{i=1}^{L},\{\text{COM}^{(i)}\}_{i=1}^{L},CLS(\cdot))$.
Here, we define $\gA(v,\gG,\vx_v)$ as the color of $v$ assigned by $\gA$.

\textit{simple AC-GNN}: The properties of aggregation, combination and classification functions are widely studied 
and many variations of these functions are proposed.
Among various function architectures, we say an AC-GNN is \textit{simple} if the aggregation and combination functions are defined as follows:
\begin{equation}
    \label{eq:agg}
    \text{AGG}^{(i)}(\sX) = \sum_{\vx \in \sX} \vx ,
   \end{equation}
   \begin{equation}
    \label{eq:simple}
    \text{COM}^{(i)}(\vx,\vy) = \sigma (\vx\mC^{(i)} + \vy\mA^{(i)}+\vb^{(i)}),
   \end{equation}
where $\mC^{(i)}$, $\mA^{(i)}$, and $\vb^{(i)}$ are trainable parameters, $\sigma$ is an activation function.

\textit{integrated AC-GNN}: The aggregation and combination functions can also be integrated such as networks explored in \cite{kipf2016semi,loukas2019graph}.
We say that such AC-GNN is \textit{integrated} when aggregation and combination functions are integrated as follows:
    \begin{align}
        \vh_u^{(i)} = \text{COM}^{(i)}(\text{AGG}^{(i)}(\{\vh_w^{(i-1)}:w\in \gN(u) \cup \{u\}\})).
       \end{align}
In {integrated} AC-GNNs, aggregation functions aggregate features from neighborhood and the node itself simultaneously, which means they treat the neighborhood information and ego-information (information from the node itself) equally.
\section{Powerful GNNs for Graph Coloring}
In this section, we focus on the question:
\textit{What kinds of designs make a GNN more/less powerful in the graph coloring problem?}
Although GNNs demonstrate their power in various tasks,
most of them even cannot beat the simplest heuristic algorithms in the coloring problem.
\begin{table}[h]
    \caption{Solved ratio of existing GNNs (GCN~\cite{kipf2016semi}, SAGE~ \cite{hamilton2017inductive}, GIN ~\cite{xu2018powerful}, GAT ~\cite{velivckovic2017graph}) and the simplest greedy algorithm on layout dataset over three runs. The node attributes are set to all-one vectors. $d$: depth.}
    \label{tab:motivation}
    \centering
    \begin{tabular}{ccc}
        \toprule
                    & $d$ = 2 & $d$= 10 \\ \hline
        GCN     & 0.55 \small{$\pm 0.01$}      & 0     \small{$\pm 0$}      \\
        SAGE   & 0  \small{$\pm 0$}        & 0      \small{$\pm 0$}     \\
        GIN    & 0.59 \small{$\pm 0.01$}      & 0.58  \small{$\pm 0.01$}      \\
        GAT      & 0   \small{$\pm 0$}       & 0 \small{$\pm 0$}          \\ \hline
        Greedy & \multicolumn{2}{c}{\textbf{0.962}}        \\
        \bottomrule
        \end{tabular}
    \end{table}
One motivating experiment is shown in \Cref{tab:motivation}, where the solved ratio is one minus the ratio between the number of conflicts and the number of edges.
For example, given a graph with 100 edges and colored with 10 conflicts, then the solved ratio is calculated by $1 - 10/100 = 0.9$.
The Greedy method colors nodes in the order of node IDs.
We observe that all tested GNNs do not work in the coloring problem.
We analyze the reason from the perspective of heterophily and homophily:
That is, linked nodes should be assigned different colors rather than the same one.
However, previous studies define the power of GNNs as the capability to map two equivalent nodes to the same embedding.
Under the heterophily, it is critical to rethink the definition of a GNN's power \textit{specifically for the coloring problem}.
After the power is formalized, the next question is: \textit{What factors may enhance or harm such a power?}
Here, we discuss the power of GNNs for graph coloring by answering the questions raised above.
We leave all proofs in Appendix due to the page limit.

\subsection{Discrimination power under heterophily}
\noindent \textbf{Q: }\textit{How to determine whether a GNN is powerful in the coloring problem}?

In the graph coloring problem, connected nodes are assigned to different colors.
Therefore, a powerful GNN should map the two connected nodes to node embeddings as differently as possible.
Intuitively, we can study the power of a GNN in the coloring problem by analyzing \textbf{\textit{its ability to assign nodes different colors.}} 
Here, we refer to the power as the \textit{discrimination power} of GNNs to differ from previous expressive power under homophily.
Formally, we define that a coloring method $f$ discriminates a node pair $(u,v)$ as follows:
\begin{mydefinition}[discriminate]
    A coloring method $f$ discriminates a node pair $(u,v)$ if $f$ assigns $u$ and $v$ different colors, i.e., $f(u,\gG,\vx_u) \neq f(v,\gG,\vx_v)$.
    \end{mydefinition}
    Following the definitions, we can answer the question above: \textit{the more powerful a GNN is, the more node pairs it will discriminate}.
    In the following context, we consider the case where the number of colors $k$ is the chromatic number $\chi$.
    Ideally, an optimal GNN should be able to discriminate all node pairs.

    Given the definition above, 
    one may try to build an optimal AC-GNN which colors all graphs without conflict through discriminating all node pairs:
    \\[8pt]
    \noindent \textbf{Q: }\textit{Can we design an AC-GNN which discriminates any node pair?}

    The following Property \ref{nonoptACGNN} refutes the existence of such a ``perfect'' AC-GNN:
    \begin{myprop}
        \label{nonoptACGNN}
    All AC-GNNs cannot discriminate any equivalent node pair.
    \end{myprop}
    One example is given in Figure \ref{fig:acgnn}, where node pair $\{c,d\}$ is equivalent. 
    Since the equivalent node pair have the same subgraph structure with the same node attribute distributions, the AC-GNN always return the same results in each layer.
    Hence, AC-GNNs are not optimal for any graph that contains these node pairs, i.e., connected and also equivalent pairs.

To avoid such a non-optimal case, we can break the equivalence between two nodes by assigning different node attributes such as random features \cite{sato2020random} or one-hot vectors \cite{barcelo2019logical}.
The solution also aligns with the conclusion in \cite{loukas2019graph}, which proves that with different attributes GNNs become significantly more powerful.
Indeed, making nodes different purposely strengthens the AC-GNN by eliminating the equivalent node pairs (although the topological equivalence is preserved).
However, the superficial methods on the node attributes cannot influence and solve the underlying defects of some specific AC-GNNs, for example, integrated AC-GNN.
In an integrated AC-GNN, the node and its neighbors are aggregated into the same multi-set,
making it more difficult for integrated AC-GNNs to discriminate the node with its neighbors.
Property \ref{nonoptHACGNN} points out that an integrated AC-GNN cannot be optimal since there always exists a set of graphs in which at least one node pair is not discriminated by the integrated AC-GNN:
\begin{myprop}
    \label{nonoptHACGNN}
        If nodes $u$ and $v$ in a graph $\gG$ are connected and share the same neighborhood except each other,
        i.e., $\gN(u) \backslash \{v\} = \gN(v)\backslash \{u\}$, then an integrated AC-GNN cannot discriminate $\{u,v\}$.
    \end{myprop}
The property points to the deficiency of integrated AC-GNN even if nodes are differentiable by their attributes.
    One example is given in Figure \ref{fig:intgnn}, where the node pair $\{c,d\}$ cannot be discriminated by any integrated AC-GNN even the node attributes are different.
\subsection{Locality}
Local methods are widely used in the combinational optimization problems such as maximum independent set (MIS) and graph coloring.
The formal definition of the local method for the coloring problem is described as follows, which is a direct rephrasing in \cite{gamarnik2014limits}:
    \begin{mydefinition}[local method \cite{gamarnik2014limits}]
        \label{def:local}
        A coloring method $f$ is r-local if it fails to discriminate any r-local equivalent node pair.
        A coloring method $f$ is local if $f$ is r-local for at least one positive integer r.
    \end{mydefinition}

Along with the study of the local methods, the upper bound of a local method for the MIS problem is investigated.
David et al.~\cite{gamarnik2014limits} gives an upper bound $1/2 + 1/(2\sqrt2)$ of an MIS produced by any local method in the random $d$-regular graph as $d\rightarrow \infty$
and \cite{rahman2017local} strengthens the bound to $1/2$. A random $d$-regular graph is a graph with $n$ nodes and the nodes in each node pair are connected with a probability $d/n$.
Starting from the upper bound of any local method for the MIS problem,
we may try to figure out:
\\[8pt]
\noindent \textbf{Q: }\textit{Whether a local method is also non-optimal in the graph coloring problem?}

The answer is given by following corollary:
\begin{mycorollary}
    \label{nonoptlocal}
        A local coloring method is non-optimal in the random d-regular tree as $d\rightarrow \infty$.
\end{mycorollary}
We finish the proof by making use of the upper bound studied in the MIS problem and bridging the connection between a local method for MIS problem and coloring problem.
Due to the localized nature of the aggregation function in GNNs, an AC-GNN with a fixed number of layers, say $L$ layers,
cannot detect the structure or information of nodes at a distance further than $L$.
Considering the non-optimality of a local coloring method stated in Corollary \ref{nonoptlocal} and the localized nature of GNNs,
we can reduce our analysis of whether there exists an optimal AC-GNN for graph coloring to whether an AC-GNN is a local coloring method?
Corollary \ref{prop:local} answers the question as yes:
    \begin{mycorollary}
        \label{prop:local}
        $L$-AC-GNN is an $L$-local coloring method and thus a local coloring method.
    \end{mycorollary}
Corollary \ref{nonoptlocal} and Corollary \ref{prop:local} directly lead to the following theorem:
\begin{mytheorem}
    \label{nonoptLocal}
 AC-GNN is not optimal, specifically for the random d-regular tree as $d\rightarrow \infty$.
\end{mytheorem}

Based on the analysis above, we can see that the locality of AC-GNN makes it infeasible to to be an optimal coloring function.
To solve the problems raised by locality of AC-GNNs, which inhibits AC-GNNs from detecting the global graph structure,
many efforts have been made to devise a global scheme such as global readout functions \cite{barcelo2019logical}, randomness \cite{you2019position,dasoulas2019coloring} and deeper networks \cite{li2019deepgcns,chen2020simple,xu2018representation}.
Among all global techniques, a deep architecture is believed to be global as long as it covers the full graph.
Given a graph with diameter $R$, a $R$-AC-GNN is able to detect the information from the whole graph.
However, it is impossible to find an AC-GNN which is able to cover all graphs: any AC-GNN is always bounded by its depth.
Then, if we cannot develop an optimal AC-GNN by simply stacking layers, does this method contribute to the discrimination power?
Formally:
\\[8pt]
\noindent \textbf{Q: }\textit{Is deeper AC-GNN more powerful in the coloring problem?}

We answer the question as yes, and give a more specific statement:
\begin{myprop}
    \label{prop:deep}
    Let $\{u,v\}$ be a node pair in any graph $\gG$, and $L$ be any positive integer. 
    If a $L$-AC-GNN discriminates $\{u,v\}$, a $L^+$-AC-GNN also discriminates it. 
\end{myprop}
A $L^+$-AC-GNN is an AC-GNN by stacking injective layers after $L$-AC-GNN (before $CLS(\cdot)$). An injective layer includes a pair of injective aggregation function and injective combination function.

\begin{figure*}[tb!]
    \subfloat[]{ \begin{tikzpicture}
\tikzstyle{every node}=[font=\small]
\begin{axis}[
width=.48\linewidth,
height=.26\linewidth,
xmin=0,
xmax=1.60,
xbar=0pt,
bar width=0.4,
yticklabels={
    GCN-2,
    GCN-10,
    GIN-2,
    GIN-10,
    SAGE-2,
    SAGE-10,
    \modelnospace-2,
    \modelnospace-10,
},
ytick={1,2,3,4,5,6,7,8},
nodes near coords,
ytick align=inside,
area legend,
legend style={
	draw=none,
	anchor=north,
	at={(1.260,1.000)},
},
reverse legend,
]
\addplot[draw=black, fill= orange!70, style={/pgf/number  format/precision=3}]
coordinates
{(0.150, 1.0) (0.7041,3) (0.786, 5)  (0.867, 6) (0.8324, 7) (0.9824, 8) (0.041, 2) (0.783,4)};

\end{axis}
\end{tikzpicture} \label{fig:sol_ratio}}
    \subfloat[]{ \begin{tikzpicture}
\tikzstyle{every node}=[font=\small]
\begin{axis}[
width=.45\linewidth,
height=.26\linewidth,
xmin=0,
xmax=1.60,
xbar=0pt,
bar width=0.4,
yticklabels={
    GCN-2,
    GCN-10,
    GIN-2,
    GIN-10,
    SAGE-2,
    SAGE-10,
    \modelnospace-2,
    \modelnospace-10,
},
ytick={1,2,3,4,5,6,7,8},
nodes near coords,
ytick align=inside,
area legend,
legend style={
	draw=none,
	anchor=north,
	at={(1.260,1.000)},
},
reverse legend,
]
\addplot[draw=black, fill= orange!70, style={/pgf/number  format/precision=3}]
coordinates
{(0.688, 1.0) (0.443,3) (0.363, 5) (1.0, 7) (0.976, 8) (0.326, 2) (0.465,4) (0.334,6)};

\end{axis}
\end{tikzpicture} \label{fig:fix_ratio}}
    \caption{(a) Solved ratio by different AC-GNN variations; (b) Fixed color ratio by different AC-GNN variations.}
\end{figure*}
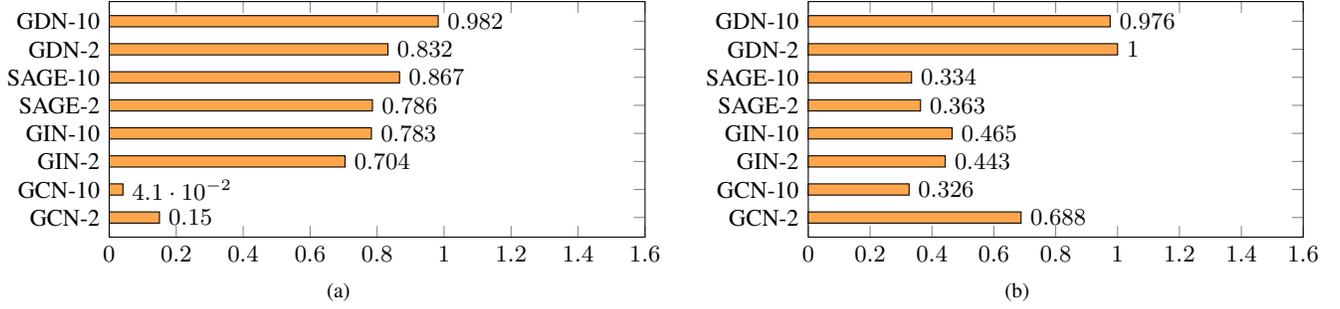

\section{Our Method}
Based on the discussions above,
we summarize a series of rules that make a GNN $\gA$ powerful in the coloring problem as follows:
\begin{enumerate}
    \item The input graph contains no equivalent node pair (Property \ref{nonoptACGNN});
    \item $\gA$ does not integrate the aggregation and combination function (Property \ref{nonoptHACGNN});
    \item $\gA$ should be as deep as possible (Property \ref{prop:deep});
    \item Layers in $\gA$ should be injective (Property \ref{prop:deep});
\end{enumerate}

With the guidance of the rules above,
we propose a very simple architecture, \textit{Graph Discrimination Network (\modelnospace)} based on simple AC-GNN.
Note that there is not solely one architecture that satisfies all the rules above.
We select \model as an example considering the balance between efficiency and performance.
We describe \model as follows:
\paragraph{Forward Computation.}
For a $k$-coloring problem, 
the node attribute is the centered probability distribution of $k$ colors and is initialized randomly to eliminate the equivalent node pairs (rule 1).

The aggregation function is the same as Equation \ref{eq:agg} (rule 2,4).
Let $\vm_v^{(i)} \in \R^{k}$ be the result returned by $\text{AGG}^{(i)}$ for the node $v$ in the $i$-th layer, the aggregation layer is organized as follows:
\begin{equation}
    \vm_v^{(i)} = \sum_{u \in \gN(v)} \vh_u^{(i-1)}.
   \end{equation}


In the combination function, we define the $\text{COM}^{(i)}$ as follows to make GDN color equivariant. For the details of color equivariance, please refer to the Appendix.

    \begin{align}
        \vh_v^{(i)} &= \vh_v^{(i-1)}  (\lambda_C^{(i)}  + \gamma_C^{(i)}(\text{\textbf{11}}^{\top}) \nonumber) \\
     & +   \vm_v^{(i)} (\lambda_A^{(i)}+\gamma_A^{(i)}(\text{\textbf{11}}^{\top})) +\beta^{(i)}\text{\textbf{1}},
    \end{align}

where $\lambda,\gamma,\beta$ are trainable scalars.
Finally, the classification function $CLS(\cdot))$ in \model is defined as an argmax function, since the final node embedding is still a probability distribution of colors.

\paragraph{Loss Function.}
Considering that the permutation of colors cannot influence the result quality,
it is not an easy job to develop a supervised training scheme since there are multiple optimal solutions.
Here, we use an un-supervised margin loss, motivated by the fact that the final node embeddings of connected nodes should be as different as possible, and formulated by:
\begin{equation}
    \label{eq:cost}
    \min \sum_{\{u,v\}\in \gE} \max \{m - d(\vh_u,\vh_v), 0\},
\end{equation}
where $\vh_u \in \R^k$ is the probability distribution obtained by $\gA$.
$d$ is the Euclidean distance between the node pair. $m$ is the pre-defined margin.

\paragraph{Preprocess \& Postprocess.}
Our method also contains preprocess and postprocess procedures, which are widely used in other coloring methods \cite{zhou2016reinforcement,li2020adaptive}.
In the preprocess part, the node with a degree less than $k$ is removed iteratively.
In the postprocess part, we iteratively detect 1) whether a color change in a single node will decrease the cost or 2) whether a swap of colors between connected nodes will decrease the cost.
We implement the two additional steps by tensor operations, which significantly boost the efficiency. 
The experiments on the two steps and detailed algorithms are shown in the Appendix D.

\paragraph{Combining with Other methods.}
As stated in Theorem 1, any AC-GNN cannot be optimal in the coloring problem. 
Therefore, we may combine with other optimal methods to obtain better quality in a sacrifice of efficiency.
Here, we propose a simple combination with ILP-based method. 
Specifically, after we obtain the color distribution of each node by GDN, 
if there is conflict(s) after $CLS(\cdot)$ , 
we can set up a threshold for the final color distributions to get a partial coloring result.
The partial result is then passed to a ILP solver to obtain the final result.
The details can be found in the Appendix.

\section{Experiments}
\label{sec:result}

\begin{table*}[!h]
    \centering
    \caption{Graph datasets information and results by different coloring methods.}
    \label{tab:dimacs}
    \resizebox{0.95\linewidth}{!}{
        \begin{tabular}{llllll|ll|ll|ll|ll}
            \toprule
            \multirow{2}{*}{\textbf{Dataset}} &
            \multirow{2}{*}{\textbf{Graph}} &
            \multirow{2}{*}{\textbf{$|\gV|$}} &
            \multirow{2}{*}{\textbf{$|\gE|$}} &
            \multirow{2}{*}{\textbf{$d\%$}} &
            \multirow{2}{*}{\textbf{$\chi(k)$}} &
            \multicolumn{2}{c|}{\textbf{GNN-GCP}} &
            \multicolumn{2}{c|}{\textbf{Tabucol}} &
            \multicolumn{2}{c|}{\textbf{HybridEA}} &
            \multicolumn{2}{c}{\textbf{GDN}} \\
            &                        &                          &                          &                      &                            & Cost          & Time         & Cost            & Time  & Cost            & Time       & Cost            & Time \\
            Layout                    & 35158                  & 641202                   & 787242                   & -                    & 3                          & 386009        & 3896         & 2392            & 82301  & {1562}            & 133285    & \textbf{1557}\small{$\pm 45$}    & 5.84 \small{$\pm 0.23$}  \\ 
            ratio                      &                        &                         &                         &                        &                          & 247.9        & 667.1         &  1.54             & 14092   & {1.00}             & 22822  & \textbf{1.0}     & \textbf{1.0}    \\ \midrule
            \multirow{3}{*}{Citation} & Cora                   & 2708                     & 5429                     & 0.15                 & 5                          & 1291          & 3.90         & 31              & 15410  & 0              & 18921    & \textbf{0} \small{$\pm 0$}     & 0.81  \small{$\pm 0.08$}  \\
                                      & Citeseer               & 3327                     & 4732                     & 0.09                 & 6                          & 1733          & 2.74         & 6               & 44700    & 0               & 24230   & \textbf{0} \small{$\pm 0$}     & 1.42 \small{$\pm 0.15$}   \\
                                      & Pubmed                 & 19717                    & 44338                    & 0.03                 & 8                          & 4393          & 4.50         & -               & $>$24h    & -               & $>$24h   & \textbf{21} \small{$\pm 4$}    & 1.41  \small{$\pm 0.12$}  \\ 
            ratio          &                 &                    &                    &                  &                    & 353.2        & 3.06         &  -             & -  &  -             & -    & \textbf{1.0}     & \textbf{1.0}    \\ \midrule
            \multirow{18}{*}{COLOR}   & jean                   & 80                       & 254                      & 8                    & 10                         & 76            & 0.06         & \textbf{0}      & 0.95   & \textbf{0}      & 0.01     & \textbf{0} \small{$\pm 1$}      & 0.13 \small{$\pm 0.02$}   \\
                                      & anna                   & 138                      & 493                      & 5                    & 11                         & 87            & 0.08         & \textbf{0}      & 3.23  & \textbf{0}      & 0.02     & \textbf{0}          \small{$\pm 0$}      & 0.17 \small{$\pm 0.03$}   \\
                                      & huck                   & 74                       & 301                      & 11                   & 11                         & 117           & 0.05         & \textbf{0}      & 0.15    & \textbf{0}      & 0.01     & \textbf{0}  \small{$\pm 0$}     & 0.06  \small{$\pm 0.02$}  \\
                                      & david                  & 87                       & 406                      & 11                   & 11                         & -             & -            & \textbf{0}      & 4.83 & \textbf{0}      & 0.01        & \textbf{0}    \small{$\pm 0$}            & 0.19 \small{$\pm 0.01$}   \\
                                      & homer                  & 561                      & 1628                     & 1                    & 13                         & 1628          & 1.09         & \textbf{0}      & 274 & \textbf{0}      & 0.06        & \textbf{0}    \small{$\pm 1$}            & 0.29 \small{$\pm 0.02$}   \\
                                      & myciel5                & 47                       & 236                      & 22                   & 6                          & 35            & 0.04         & \textbf{0}      & 0.20 & \textbf{0}      & 0.01       & \textbf{0} \small{$\pm 0$}      & 0.12 \small{$\pm 0.01$}   \\
                                      & myciel6                & 95                       & 755                      & 17                   & 7                          & 94            & 4.33         & \textbf{0}      & 0.79 & \textbf{0}      & 0.01       & \textbf{0} \small{$\pm 0$}      & 0.21 \small{$\pm 0.02$}   \\
                                      & games120               & 120                      & 638                      & 9                    & 9                          & 301           & 0.07         & \textbf{0}      & 0.93  & \textbf{0}      & 0.01       & \textbf{0} \small{$\pm 1$}      & 0.08 \small{$\pm 0.01$}   \\
                                      & Mug88\_1               & 88                       & 146                      & 4                    & 3                          & 146           & 0.33         & \textbf{0}      & 0.12  & \textbf{0}      & 0.01       & \textbf{0} \small{$\pm 0$}      & 0.01 \small{$\pm 0$}   \\
                                      & 1-Insertions\_4        & 67                       & 232                      & 10                   & 2                          & 42            & 0.05         & \textbf{0}      & 0.16  & \textbf{0}      & 0.01      & \textbf{0}  \small{$\pm 0$}     & 0.07 \small{$\pm 0$}   \\
                                      & 2-Insertions\_4        & 212                      & 1621                     & 7                    & 4                          & 360           & 0.09         & \textbf{1}      & 255  & \textbf{1}      & 101.1       & \textbf{1}     \small{$\pm 0$}           & 60.08  \small{$\pm 0.01$}  \\
                                      & Queen5\_5              & 25                       & 160                      & 53                   & 5                          & 37            & 0.03         & \textbf{0}      & 0.13   & \textbf{0}      & 0.07    & \textbf{0} \small{$\pm 0$}      & 0.05 \small{$\pm 0.01$}   \\
                                      & Queen6\_6              & 36                       & 290                      & 46                   & 6                          & 290           & 0.38         & \textbf{0}      & 4.93  & \textbf{0}      & 1.89      & \textbf{0}   \small{$\pm 0$}             & 0.08 \small{$\pm 0$}   \\
                                      & Queen7\_7              & 49                       & 476                      & 40                   & 7                          & 126           & 0.04         & {10}     & 36.9   & \textbf{9}     & 51.8      & \textbf{9}   \small{$\pm 1$}            & 0.38  \small{$\pm 0.08$}  \\
                                      & Queen8\_8              & 64                       & 728                      & 36                   & 8                          & 188           & 0.05         & 8               & 61.3 & 5               & 74.1       & \textbf{2} \small{$\pm 0$}      & 0.13  \small{$\pm 0.03$}  \\
                                      & Queen9\_9              & 81                       & 1056                     & 33                   & 9                          & 296           & 0.07         & \textbf{5}      & 97.8  & 6     & 126.9        & {6}         \small{$\pm 1$}      & 0.09 \small{$\pm 0.01$}   \\
                                      & Queen8\_12             & 96                       & 1368                     & 30                   & 12                         & 260           & 0.10         & 10              & 139  & 3              &  92.9       & \textbf{0}  \small{$\pm 0$}     & 0.58 \small{$\pm 0.09$}   \\
                                      & Queen11\_11            & 121                      & 3960                     & 55                   & 11                         & 396           & 0.10         & 33              & 213  & 22              & 141.3       & \textbf{21} \small{$\pm 3$}     & 0.07 \small{$\pm 0.01$}   \\
                                      & Queen13\_13            & 169                      & 6656                     & 47                   & 13                         & 728           & 0.20         & {42}     & 401  & 37     & 213       & \textbf{33} \small{$\pm 2$}             & 2.38  \small{$\pm 0.32$}   \\ 
                                      ratio          &                 &                    &                    &                  &                 & -        & -         &  {1.51}             & 22.63  &  {1.15}             & 12.10   & \textbf{1.0}     & \textbf{1.0}    \\ \bottomrule
            
        \end{tabular}
    }
\end{table*}

\subsection{Experimental setup}
Detailed settings, dataset introduction, more experiments and analysis are shown in the Appendix.
We evaluate our models and baselines on three datasets here, the basic information on these datasets are shown in \Cref{tab:dimacs}, where column $\chi(k)$ is the chromatic number except layout dataset, which is set to 3 in the real-world circuit design.

We mainly compare our models with three previous works (the results of other methods such as ILP and simple heuristics can be found in the Appendix):
(1) GNN-GCP \cite{lemos2019graph}, combing GNN, RNN, and MLP to obtain the node embedding and using a k-means method to color the node. 
We obtain models from the author and directly obtain the results.
(2) Tabucol \cite{hertz1987using}, a well-known heuristic algorithm using Tabu search.
We follow the original setting with an iteration limit of 1000 (or the time limit of 24 hours) and the number of uncolored node pairs is returned if the algorithm fails to find a perfect coloring assignment within the limit.
(3) HybridEA \cite{galinier1999hybrid}, the state-of-the-art evolutionary algorithm for the coloring problem. 
We also compare different variants of AC-GNN in the previous works:
GCN \cite{kipf2016semi}, GIN \cite{xu2018powerful}, and GraphSAGE \cite{hamilton2017inductive}.
All AC-GNN variations are only tested in the layout dataset since AC-GNN variations require a fixed number of colors to make output shape keep the same.
To make the comparison fair, we use the same CPU to test all methods.

\subsection{Comparison with other AC-GNN variations}
The results are shown in \Cref{fig:sol_ratio}.
“GDN-$k$" represents \model with a depth of $k$.
According to the results, we observe the following: 
(1) GCN, the most representative integrated AC-GNN, is much worse than other AC-GNNs, which demonstrates our rule 1.
(2) Most AC-GNNs benefit from a deeper network, which aligns with our rule 3.
(3) Although some non-integrated AC-GNN achieve an acceptable solved ratio, 
our method is still far better than other AC-GNNs.

We also validate the color equivariance of models by simulating the pre-color constraint in the layout decomposition problem.
For each instance, we randomly select one node and set its color by changing the node attribute, known as the color distribution.
We measure the color equivariant capability by checking the fixed color ratio, 
defined as the ratio between the number of successfully fixed graphs and the number of total graphs.
A successfully fixed graph is the graph whose selected node is colored as expected with the pre-assigned one .
From the results shown in \Cref{fig:fix_ratio}, we can see that the fixed color ratio of our \model is much higher than other variations, matching with our analysis. 

\subsection{Comparison with other graph coloring methods}
The comparison with other graph coloring methods is conducted on all collected datasets. 
We combine with ILP for the complex dataset, i.e., the COLOR dataset.
The results are shown in \Cref{tab:dimacs}, where $k$ is the number of available colors and cost is the number of conflicts in the coloring result.
GNN-GCP gives ``-'' if it fails to find a chromatic number prediction.
Tabucol and HybridEA give ``-'' if they fails to color the graph within 24 hours.
According to the results, we observe the following:
(1) Our method outperforms the state-of-the-art algorithms with a better result quality and 10$\times$ speedup.
(2) Our method is more advantageous for complex and large graphs (Citation and Queen), which are more beneficial for industrial demand.



\section{Conclusion}

In this paper, we established theoretical foundations for reasoning about the discrimination power of GNNs for the graph coloring problem.
We identified the node pairs that a popular class of GNNs, AC-GNNs fail to discriminate and 
gave conditions on how an AC-GNN can be more discriminatively powerful. 
Moreover, we built the connection between the locality study in graph theory and the local property of AC-GNNs, 
and proved the non-optimality of AC-GNN due to the locality.
Furthermore, we analyzed the color equivariance in the graph coloring problem and proposed a scheme to make AC-GNN color equivariant.
Combining all the analysis above,
we designed a simple variation of AC-GNN for the graph coloring problem, which proves to be discriminatively powerful and color equivariant.
To complete the picture, it would be interesting to analyze the global terms for enhancing the discrimination power of GNNs.

\bibliographystyle{IEEEtran}
\bibliography{ref/Top,ref/reference}

\begin{thebibliography}{10}
\providecommand{\url}[1]{#1}
\csname url@samestyle\endcsname
\providecommand{\newblock}{\relax}
\providecommand{\bibinfo}[2]{#2}
\providecommand{\BIBentrySTDinterwordspacing}{\spaceskip=0pt\relax}
\providecommand{\BIBentryALTinterwordstretchfactor}{4}
\providecommand{\BIBentryALTinterwordspacing}{\spaceskip=\fontdimen2\font plus
\BIBentryALTinterwordstretchfactor\fontdimen3\font minus
  \fontdimen4\font\relax}
\providecommand{\BIBforeignlanguage}[2]{{%
\expandafter\ifx\csname l@#1\endcsname\relax
\typeout{** WARNING: IEEEtran.bst: No hyphenation pattern has been}%
\typeout{** loaded for the language `#1'. Using the pattern for}%
\typeout{** the default language instead.}%
\else
\language=\csname l@#1\endcsname
\fi
#2}}
\providecommand{\BIBdecl}{\relax}
\BIBdecl

\bibitem{hamilton2017representation}
W.~L. Hamilton, R.~Ying, and J.~Leskovec, ``Representation learning on graphs:
  Methods and applications,'' \emph{arXiv preprint arXiv:1709.05584}, 2017.

\bibitem{barcelo2019logical}
P.~Barcel{\'o}, E.~V. Kostylev, M.~Monet, J.~P{\'e}rez, J.~Reutter, and J.~P.
  Silva, ``The logical expressiveness of graph neural networks,'' in
  \emph{International Conference on Learning Representations (ICLR)}, 2019.

\bibitem{loukas2019graph}
A.~Loukas, ``What graph neural networks cannot learn: depth vs width,''
  \emph{arXiv preprint arXiv:1907.03199}, 2019.

\bibitem{xu2018powerful}
K.~Xu, W.~Hu, J.~Leskovec, and S.~Jegelka, ``How powerful are graph neural
  networks?'' \emph{arXiv preprint arXiv:1810.00826}, 2018.

\bibitem{gama2020stability}
F.~Gama, J.~Bruna, and A.~Ribeiro, ``Stability properties of graph neural
  networks,'' \emph{IEEE Transactions on Signal Processing}, vol.~68, pp.
  5680--5695, 2020.

\bibitem{loukas2020hard}
A.~Loukas, ``How hard is to distinguish graphs with graph neural networks?''
  Tech. Rep., 2020.

\bibitem{du2019graph}
S.~S. Du, K.~Hou, B.~P{\'o}czos, R.~Salakhutdinov, R.~Wang, and K.~Xu, ``Graph
  neural tangent kernel: Fusing graph neural networks with graph kernels,''
  \emph{arXiv preprint arXiv:1905.13192}, 2019.

\bibitem{rahman2017local}
M.~Rahman, B.~Virag \emph{et~al.}, ``Local algorithms for independent sets are
  half-optimal,'' \emph{The Annals of Probability}, vol.~45, no.~3, pp.
  1543--1577, 2017.

\bibitem{gamarnik2014limits}
D.~Gamarnik and M.~Sudan, ``Limits of local algorithms over sparse random
  graphs,'' in \emph{Proceedings on Innovations in theoretical computer
  science}, 2014, pp. 369--376.

\bibitem{kipf2016semi}
T.~N. Kipf and M.~Welling, ``Semi-supervised classification with graph
  convolutional networks,'' \emph{arXiv preprint arXiv:1609.02907}, 2016.

\bibitem{hamilton2017inductive}
W.~Hamilton, Z.~Ying, and J.~Leskovec, ``Inductive representation learning on
  large graphs,'' in \emph{Annual Conference on Neural Information Processing
  Systems (NIPS)}, 2017, pp. 1024--1034.

\bibitem{velivckovic2017graph}
P.~Veli{\v{c}}kovi{\'c}, G.~Cucurull, A.~Casanova, A.~Romero, P.~Lio, and
  Y.~Bengio, ``Graph attention networks,'' \emph{arXiv preprint
  arXiv:1710.10903}, 2017.

\bibitem{sato2020random}
R.~Sato, M.~Yamada, and H.~Kashima, ``Random features strengthen graph neural
  networks,'' \emph{arXiv preprint arXiv:2002.03155}, 2020.

\bibitem{you2019position}
J.~You, R.~Ying, and J.~Leskovec, ``Position-aware graph neural networks,''
  \emph{arXiv preprint arXiv:1906.04817}, 2019.

\bibitem{dasoulas2019coloring}
G.~Dasoulas, L.~D. Santos, K.~Scaman, and A.~Virmaux, ``Coloring graph neural
  networks for node disambiguation,'' \emph{arXiv preprint arXiv:1912.06058},
  2019.

\bibitem{li2019deepgcns}
G.~Li, M.~Muller, A.~Thabet, and B.~Ghanem, ``{DeepGCNs}: Can {GCNs} go as deep
  as {CNNs}?'' in \emph{IEEE International Conference on Computer Vision
  (ICCV)}, 2019, pp. 9267--9276.

\bibitem{chen2020simple}
M.~Chen, Z.~Wei, Z.~Huang, B.~Ding, and Y.~Li, ``Simple and deep graph
  convolutional networks,'' in \emph{International Conference on Machine
  Learning}.\hskip 1em plus 0.5em minus 0.4em\relax PMLR, 2020, pp. 1725--1735.

\bibitem{xu2018representation}
K.~Xu, C.~Li, Y.~Tian, T.~Sonobe, K.-i. Kawarabayashi, and S.~Jegelka,
  ``Representation learning on graphs with jumping knowledge networks,'' in
  \emph{International Conference on Machine Learning}.\hskip 1em plus 0.5em
  minus 0.4em\relax PMLR, 2018, pp. 5453--5462.

\bibitem{zhou2016reinforcement}
Y.~Zhou, J.-K. Hao, and B.~Duval, ``Reinforcement learning based local search
  for grouping problems: A case study on graph coloring,'' \emph{Expert Systems
  with Applications}, vol.~64, pp. 412--422, 2016.

\bibitem{li2020adaptive}
W.~Li, J.~Xia, Y.~Ma, J.~Li, Y.~Liny, and B.~Yu, ``Adaptive layout
  decomposition with graph embedding neural networks,'' in \emph{2020 57th
  ACM/IEEE Design Automation Conference (DAC)}.\hskip 1em plus 0.5em minus
  0.4em\relax IEEE, 2020, pp. 1--6.

\bibitem{lemos2019graph}
H.~Lemos, M.~Prates, P.~Avelar, and L.~Lamb, ``Graph colouring meets deep
  learning: Effective graph neural network models for combinatorial problems,''
  in \emph{IEEE International Conference on Tools with Artificial Intelligence
  (ICTAI)}.\hskip 1em plus 0.5em minus 0.4em\relax IEEE, 2019, pp. 879--885.

\bibitem{hertz1987using}
A.~Hertz and D.~de~Werra, ``Using tabu search techniques for graph coloring,''
  \emph{Computing}, vol.~39, no.~4, pp. 345--351, 1987.

\bibitem{galinier1999hybrid}
P.~Galinier and J.-K. Hao, ``Hybrid evolutionary algorithms for graph
  coloring,'' \emph{Journal of combinatorial optimization}, vol.~3, no.~4, pp.
  379--397, 1999.

\end{thebibliography}

\end{document}